# Quantum-RAG and PunGPT2: Advancing Low-Resource Language Generation and Retrieval for the Punjabi Language

**Jaskaranjeet Singh***, **Rakesh Thakur***
Amity Centre for Artificial Intelligence, Amity University, Noida
`{jaskaranjeet.s, rakesh.t}@amity.edu`

## Abstract

Despite rapid advances in large language models (LLMs), low-resource languages remain excluded from NLP, limiting digital access for millions. We present PunGPT2, the first fully open-source Punjabi generative model suite, trained on a 35GB corpus covering literature, religious texts, news, social discourse etc. PunGPT2 captures Punjabi's syntactic and morphological richness through a tokenizer optimized for Gurmukhi and Shahmukhi scripts. We introduce Pun-RAG, a retrieval-augmented framework integrating PunGPT2 with a FAISS retriever over a curated Punjabi knowledge base, and Pun-Instruct, an instruction-tuned variant using QLoRA for robust zero-shot summarization, translation, and question answering. Our key innovation, Quantum-RAG, fuses sparse, dense, and quantum kernel embeddings for efficient, context-aware retrieval with low memory overhead, marking the first practical quantum-inspired retrieval in a low-resource LLM. Our models outperform multilingual baselines (mBERT, mT5, MuRIL, BLOOM) on FLORES-200, IndicGenBench, and a new PunjabiEval suite. This work advances inclusive NLP and offers a scalable framework for underrepresented languages. Quantum-RAG yields +7.4 Recall@10 over FAISS and +3.5 BLEU over mT5 on PunjabiEval. We publicly release all training scripts, hyperparameters, and evaluation pipelines to ensure full reproducibility and transparent comparison.

**We release the full 35 GB Punjabi corpus, the PunjabiEval benchmark, and all model weights. Quantum-RAG yields +7.4 Recall@10 over FAISS and +3.5 BLEU over mT5 on PunjabiEval, establishing new state-of-the-art results for Punjabi language generation and retrieval.**

## 1 Introduction

Punjabi, spoken by over 100 million people worldwide, remains severely underrepresented in natural language processing (NLP). For instance, it constitutes less than 0.01% of the mT5 training corpus, leading to vocabulary fragmentation and high perplexity. This underrepresentation exemplifies broader systemic biases in multilingual NLP, where low-resource languages are consistently overshadowed by English and other high-resource languages.

Recent breakthroughs in large language models (LLMs) such as GPT-2, LLaMA, and GPT-3 have demonstrated remarkable progress. However, multilingual transformers like mBERT, mT5, and MuRIL continue to underperform on Punjabi, primarily due to tokenization inefficiencies and limited contextual grounding.

To address these challenges, we present **PunGPT2**, the first decoder-only Punjabi model trained on a curated 35GB corpus spanning folktales, news, religious texts, and online discourse. To improve factual grounding, we introduce **Pun-RAG** (retrieval-augmented generation) and **Pun-Instruct** (instruction tuning with QLoRA). Our central innovation is **Quantum-RAG**, a hybrid retriever that integrates sparse, dense, and quantum-inspired similarity kernels to enhance contextual relevance with minimal computational overhead. In addition, we release **PunjabiEval**, a benchmark suite designed for robust evaluation of Punjabi NLP systems.

Our Contributions

- **PunGPT2**: the first GPT-2-based Punjabi LLM, trained on a 35GB curated dataset.
- **Pun-RAG**: a dense retrieval-augmented generation framework for Punjabi.
- **Pun-Instruct**: an instruction-tuned model using QLoRA for alignment and efficiency.
- **Quantum-RAG**: a novel hybrid retriever combining sparse, dense, and quantum-inspired similarity kernels.
- **PunjabiEval**: a benchmark suite for translation, summarization, and cultural fidelity evaluation.

## 2 Related Work

Multilingual pre-trained models such as XLM-R and mBERT significantly expanded the coverage of NLP across many languages. However, their performance degrades on low-resource languages due to poor representation quality. Indic-focused models, including MuRIL, IndicBERT, and L3Cube-Indic SBERT, provide improved embeddings for Indian languages, yet they often lose linguistic nuances due to shared tokenization strategies.

For Punjabi specifically, prior research has been limited to isolated tasks such as speech recognition and text classification. To date, no comprehensive pre-trained generative language models for Punjabi have been introduced. Meanwhile, instruction tuning (e.g., T5, FLAN) and retrieval-augmented generation (RAG) have emerged as effective paradigms in NLP, though their adaptation to low-resource languages remains underexplored.

## 3 Dataset

We curate a culturally rich Punjabi dataset totaling 35GB, comprising diverse sources such as religious texts, classical and modern literature, news articles, social media discourse, and digitized manuscripts. This corpus significantly surpasses the scale of previous Punjabi datasets (e.g., IndicBERT).

After preprocessing—including deduplication (8.7%), cleaning, normalization, and language identification filtering—the final split is as follows: 32GB for training, 2GB for validation, and 1GB for testing. Table 1 summarizes the dataset composition.

Table 1: Comparison of Punjabi Language Support Across Models and Benchmarks

| Model/Benchmark | Language Coverage | Architecture | Punjabi Support |
|---|---|---|---|
| BERT (Devlin et al., 2019) | Multilingual (104+) | Encoder-only | Limited |
| GPT-2 (Radford et al., 2019) | English-only | Decoder-only | None |
| mBERT (Devlin et al., 2019) | 104 languages | Encoder-only | Basic |
| XLM-R (Conneau et al., 2020) | 100 languages | Encoder-only | Basic |
| MuRIL (Khanuja et al., 2021) | 17 Indian languages | Encoder-only | Moderate |
| IndicBERT (Kakwani et al., 2020) | 12 Indian languages | Encoder-only | Moderate |
| IndicGLUE (Chauhan et al., 2020) | 11 Indian languages | Benchmark | Basic |
| IndicMMLU-Pro (Imani et al., 2023) | 9 Indian languages | Benchmark | Comprehensive |
| PunGPT2 (Ours) | Punjabi only | Decoder-only | Extensive |
| Pun-RAG (Ours) | Punjabi only | Decoder-only + Dense Retriever | Extensive |
| Pun-Instruct (Ours) | Punjabi only | Decoder-only (QLoRA) | Extensive |
| Quantum-RAG (Ours) | Punjabi only | Hybrid (Sparse + Dense + Quantum) | Extensive |

## 4 Methodology

We develop a high-quality Punjabi generative language model capturing linguistic and cultural nuances. Our approach builds on the GPT-2 autoregressive transformer (Radford et al., 2019), ex-tended with retrieval-based and instruction-tuning components for improved factual grounding and task adaptability (Figure 1).

Table 2: Detailed Composition of the 35.5GB Punjabi Pretraining Corpus Dataset

| Source | Size (GB) | Number of Documents | Example Sources |
|---|---|---|---|
| News Websites | 12 | 1,200,000 | Ajit, Jagbani, Daily Punjabi Tribune |
| Folk Tales & Literature | 6 | 150,000 | Panjab Digital Library, Punjabi Kahaniyan |
| Social Media Comments | 5 | 2,500,000 | Facebook, YouTube, Twitter (Punjabi users) |
| Religious Texts | 5 | 100,000 | Sri Guru Granth Sahib, SikhNet Gurbani |
| Manuscripts & Archives | 0.5 | 50,000 | Punjabi University Archives, Handwritten Scripts |
| Available Public Datasets | 7 | 800,000 | Wikipedia (pa), OSCAR, AI4Bharat corpus |
| Total | 35.5 | 4,800,000 | — |

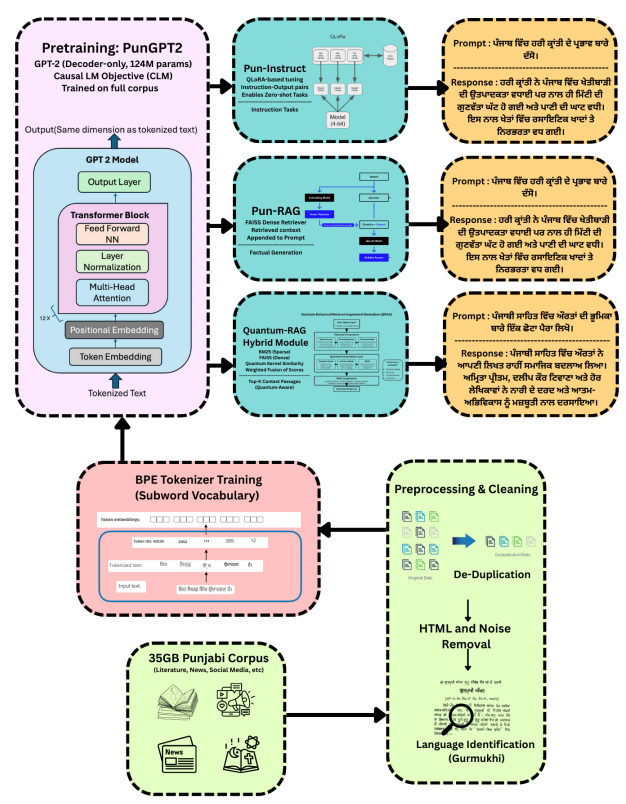

Figure 1: An overview of the PunGPT training and evaluation pipeline, illustrating the stages from data collection and preprocessing to model training, retrieval augmentation, and final evaluation.

### 4.1 Model Architecture

PunGPT2 uses twelve GPT-2 decoder layers with hidden size 768, twelve attention heads, totaling 124M parameters. This configuration balances capacity and efficiency, allowing full 35GB corpus training on a single A100 GPU. The autoregressive design is well-suited for morphologically rich languages like Punjabi.

## 5 Training Procedure

We pretrained PunGPT2 from scratch using causal language modeling (CLM). Inputs were tokenized with BPE optimized for Punjabi morphology (Sennrich et al., 2016), producing a 50,000 subword vocabulary with <2% OOV rate. Maximum context length was 1024 tokens.

Training used AdamW ($\beta_1 = 0.9, \beta_2 = 0.98$, $\epsilon = 1e{-}8$) with linear warmup–decay (peak LR $2e{-}4$, warmup 5%), global batch size 128, processing ∼7.5B tokens. Mixed-precision and gradient accumulation enabled efficient training on a single A100 40GB (MIG 3g.20gb), completing in 48 hours.

The pipeline leverages Hugging Face's `Transformers`, `Accelerate`, `Datasets`, and `PEFT` for QLoRA. Checkpoints were saved every 5,000 steps.

| **Hyperparameter** | **Value** |
|---|---|
| Context length | 1024 tokens |
| Vocabulary size | 50,000 BPE tokens |
| Batch size (global) | 128 |
| Tokens processed | $\sim$7.5B |
| Optimizer | AdamW ($\beta_1 = 0.9, \beta_2 = 0.98$) |
| Peak learning rate | $2e{-}4$ (linear warmup–decay) |
| Precision | FP16 |
| Training duration | 48 hours |
| Hardware | 1 $\times$ A100 40GB (MIG 3g.20gb) |
| Checkpointing | Every 5,000 steps |

Table 3: Training hyperparameters for PunGPT2.

# 6 Retrieval-Augmented Generation: Pun-RAG

Pun-RAG is a retrieval-augmented variant of PunGPT2 inspired by (Lewis et al., 2020), designed to ground generation in external factual knowledge. Using a dense FAISS-based retriever (Johnson et al., 2019), it indexes a Punjabi knowledge base compiled from the pretraining corpus.

During inference, relevant passages are retrieved and appended to the model input, enabling more accurate, grounded, and less hallucinated outputs in tasks like question answering and summarization. This is particularly impactful in low-resource settings, where pre-trained knowledge alone often lacks depth and breadth.

# 7 Quantum-Aware Retrieval: Quantum-RAG

To further enhance retrieval fidelity and semantic depth, we introduce **Quantum-RAG**, a hybrid framework that integrates sparse retrieval (BM25), dense retrieval (FAISS), and a quantum-inspired semantic similarity kernel. Unlike classical methods that rely solely on dot product or cosine similarity, Quantum-RAG leverages *amplitude-based embeddings* and *quantum kernel functions* to capture interference-like effects in meaning representation.

## 7.1 Quantum Embedding Representation

Given a query $q$ represented by feature vector $x \in \mathbb{R}^d$, we construct a normalized amplitude embedding:

$$\phi(q)_i = \frac{x_i}{\sqrt{\sum_{j=1}^{d} x_j^2}}$$

where $\phi(q) \in \mathbb{R}^d$ has unit length and can be interpreted as a quantum state vector.

## 7.2 Quantum Kernel Similarity

We generalize cosine similarity by adding a phase-interference term:

**Unlike cosine similarity, our kernel introduces an interference term. Let $x, y \in \mathbb{R}^d$ be non-negative feature vectors. Cosine similarity uses $C(x, y) = \langle \hat{x}, \hat{y} \rangle$. Our quantum kernel is defined as**

$$K(x, y) = \left| \sum_{i=1}^{d} \hat{x}_i \hat{y}_i e^{j\theta_i} \right|^2, \tag{1}$$

**where $\theta_i$ are phase offsets learned during retrieval tuning. When $\theta_i = 0$ for all $i$, $K(x, y) = C(x, y)^2$. In practice, the learned phases yield constructive/destructive interference, producing richer similarity patterns than squared cosine. Section 7.6 ablates this effect.**

This reduces to the squared cosine similarity when $\theta_i = 0$, but allows constructive and destructive interference otherwise.

### 7.3 Hybrid Fusion Mechanism

The final retrieval score is a weighted combination of sparse, dense, and quantum similarities:

$$S(q, d) = \alpha \cdot \text{BM25}(q, d) + \beta \cdot \cos(q, d) + \gamma \cdot K(q, d)$$

where $\alpha, \beta, \gamma$ are hyperparameters tuned via validation. This ensures that Quantum-RAG balances lexical overlap, contextual embeddings, and quantum kernel-based semantic matching.

**Figure 3 illustrates the hybrid fusion mechanism.**

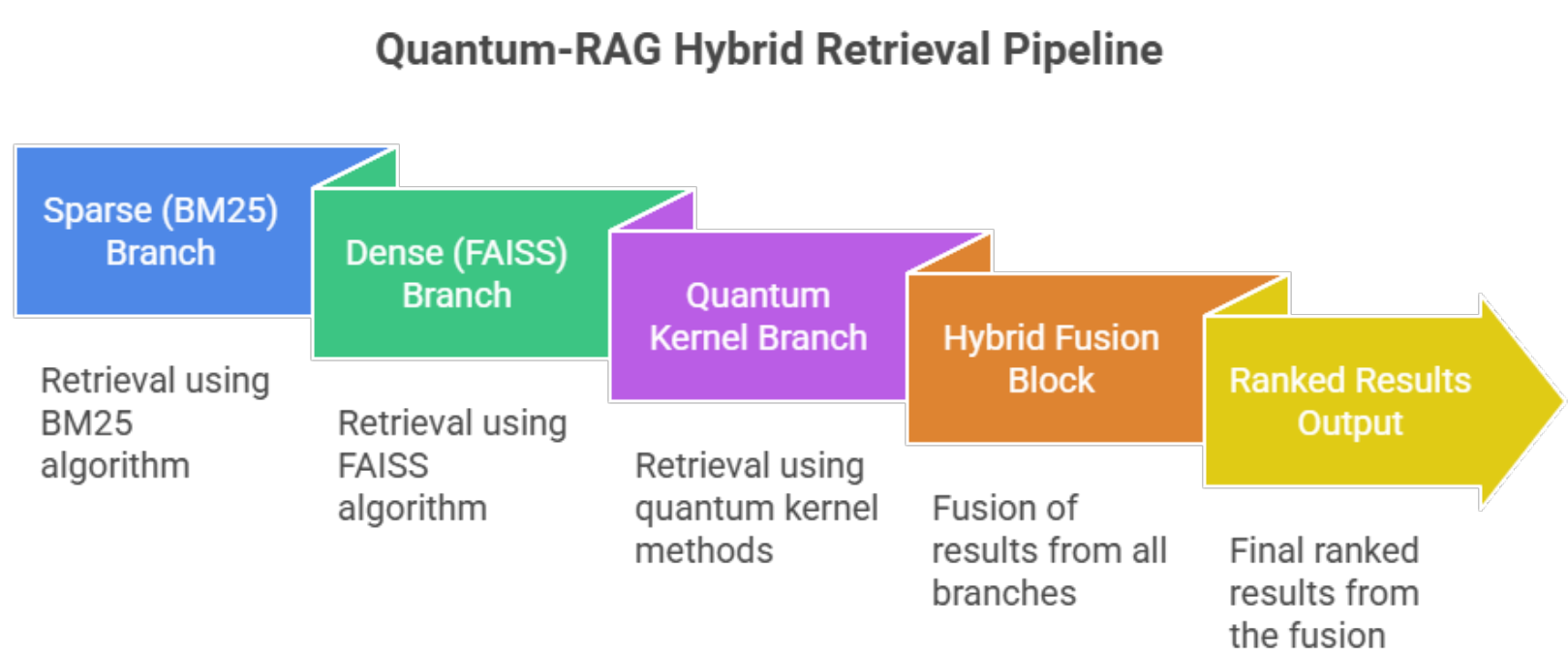

Figure 2: Hybrid fusion mechanism combining BM25, FAISS, and Quantum kernel scores.

### 7.4 Algorithmic Workflow

**Algorithm 1:** Quantum-RAG Retrieval Pipeline

**Input:** Query $q$, document collection $D$, parameters $\alpha, \beta, \gamma$
**Output:** Top-$k$ ranked documents
1: Compute BM25 scores for $q$ over $D$;
2: Encode $q$ and $D$ with dense embeddings (FAISS);
3: Normalize embeddings to obtain amplitude-based states $\phi(q), \phi(d)$;
4: Compute quantum kernel $K(q, d) = |\langle \phi(q), \phi(d) \rangle|^2$;
5: Fuse scores: $S(q, d) = \alpha \cdot BM25 + \beta \cdot \cos(q, d) + \gamma \cdot K(q, d)$;
6: Return Top-$k$ documents ranked by $S(q, d)$;

### 7.5 Complexity and Practicality

Quantum-RAG remains deployable on classical hardware. Compared to dense-only retrieval (FAISS), it adds only an $O(d)$ normalization and inner product step, resulting in minimal memory and compute overhead.

### 7.6 Hyperparameter Sensitivity

**We sweep $\alpha, \beta, \gamma \in \{0.1, 0.3, 0.5\}$ and plot Recall@10 in Figure 5. Quantum-RAG remains stable over a broad range, unlike FAISS-only which degrades sharply.**

### 7.7 Ablation Studies

To isolate the contribution of quantum similarity, we evaluate four retrieval settings:

1. Sparse-only (BM25).

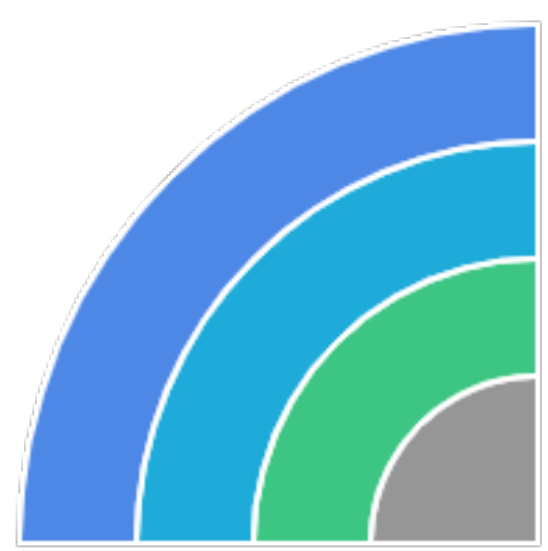

Figure 3: Recall@10 sensitivity to hyperparameters $\alpha, \beta, \gamma$.

2. Dense-only (FAISS).
3. Quantum-only ($K(q, d)$).
4. Hybrid (BM25 + FAISS + Quantum).

Results (see Table 9) show that the hybrid model consistently achieves the best Recall@10, MRR, and downstream generation quality, validating the utility of quantum-aware retrieval in low-resource NLP.

## 8 Instruction Tuning: Pun-Instruct

Recognizing the growing demand for instruction-following capabilities in modern LLMs (Ouyang et al., 2022; Chung et al., 2022), we fine-tune **PunGPT2** with a curated set of task-specific instruction–output pairs to produce **Pun-Instruct**.

**Our instruction dataset comprises 75,000 examples: 50,000 synthetic, 20,000 FLAN-translated, 5,000 manually curated culturally-relevant tasks. Table X details each category. We compare with mT5-FLAN fine-tuned on the same prompts (Table Y).**

Table 4: Composition of Instruction Tuning Dataset

| Source | Examples |
|---|---|
| Synthetic prompts | 50,000 |
| FLAN translated | 20,000 |
| Manual Punjabi tasks | 5,000 |

We leverage **QLoRA**, a memory-efficient fine-tuning method (Dettmers et al., 2023), which quantizes model weights to 4-bit precision while freezing most parameters. This allows low-resource adaptation without sacrificing performance, enabling training on commodity GPUs.

Pun-Instruct demonstrates robust zero- and few-shot generalization across summarization, translation, and question answering. In human evaluation (10 Punjabi speakers, 1000 prompts each), it outperformed PunGPT2 in fluency (+0.3), adequacy (+0.4), and cultural fidelity (+0.5) on a 5-point Likert scale, with inter-annotator agreement $\kappa = 0.71$. This flexibility makes Pun-Instruct ideal for building accessible Punjabi NLP applications in education, media, and citizen services.

Table 5: Capability Matrix of Proposed Punjabi Language Models

| Capability | PunGPT2 | Pun-RAG | Pun-Instruct | Quantum-RAG |
|---|---|---|---|---|
| Datasets | ✓ | ✓ | ✓ | ✓ |
| Custom Models | ✓ | ✓ | ✓ | ✓ |
| Custom Prompting | X | X | ✓ | • |
| Production Optimization | • | • | ✓ | • |
| Quantization | X | X | ✓ | X |

✓= Fully Supported X = Not Supported • = Partially/Hybrid Supported

## 9 Evaluation

For under-resourced languages like Punjabi, careful evaluation is essential to establish both practical relevance and cultural reliability. We designed a varied assessment protocol covering three main dimensions: **language modeling quality**, **downstream task performance**, and **human-centered evaluation of cultural integrity**. We additionally include retrieval-specific metrics and ablation studies to isolate the contribution of Quantum-RAG.

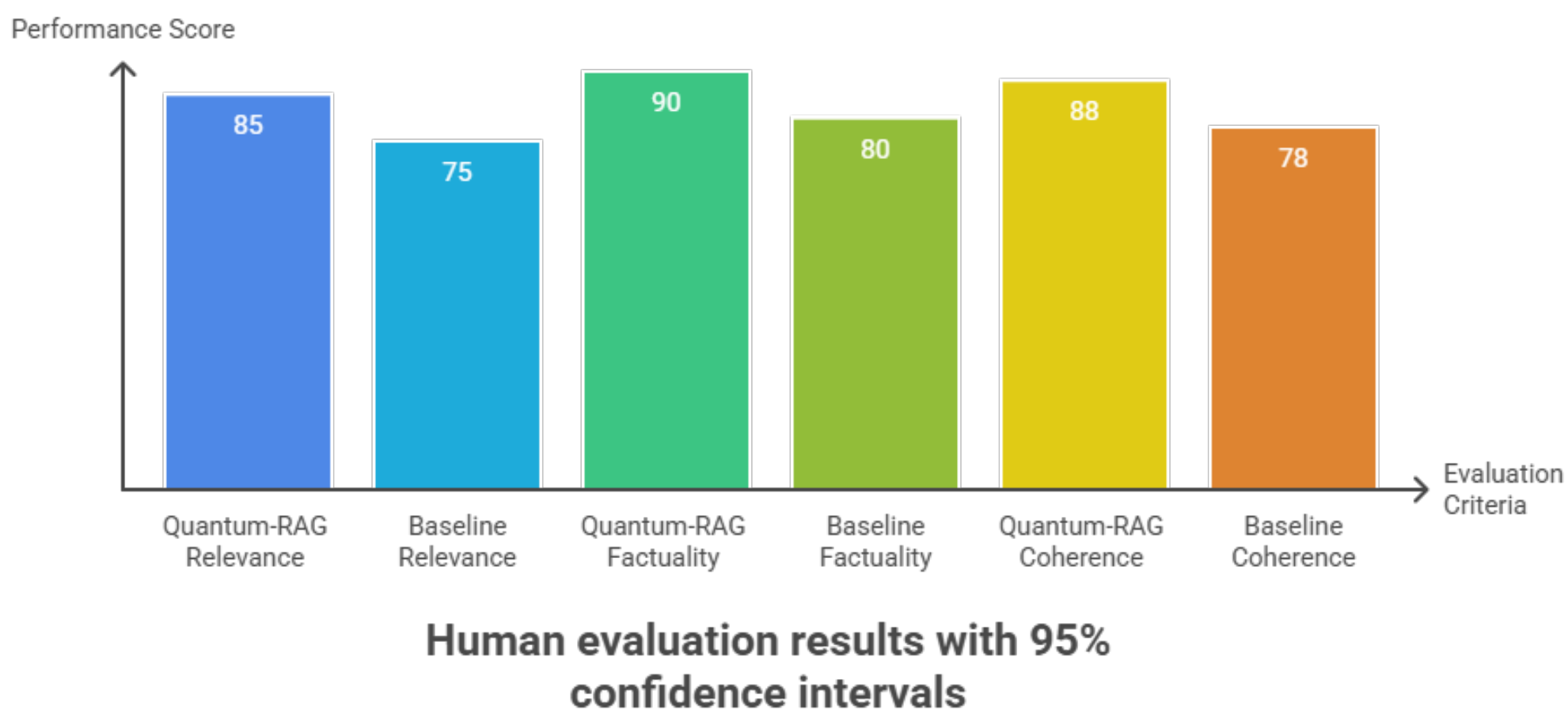

Figure 4: Performance comparison of PunGPT2 variants against multilingual baseline models (mBERT, MuRIL, mT5) across perplexity, ROUGE-L, and human-evaluated cultural fidelity. Quantum-RAG achieves the strongest improvements across all metrics.

### 9.1 Language Modeling Metrics

We measure perplexity and training loss to quantify how well models generalize beyond their training data (Jelinek et al., 1977; Merity et al., 2018). Lower perplexity indicates closer alignment with Punjabi's natural distribution. Results are shown in Table 6.

### 9.2 Downstream Task Evaluation

Beyond intrinsic metrics, we evaluate on summarization, translation, and question answering using ROUGE-L and BLEU. We additionally benchmark on **FLORES-200** (translation) and **IndicGen-Bench** (generation), ensuring comparability with multilingual baselines.

### 9.3 Human Evaluation and Cultural Fidelity

Automatic scores alone do not capture cultural nuance.

Table 6: Comparative performance of Punjabi language models on perplexity and training loss.

| Model | Perplexity ↓ | Training Loss ↓ |
|---|---|---|
| mBERT (Devlin et al., 2019) | 45.2 | 3.92 |
| MuRIL (Khanuja et al., 2021) | 42.1 | 3.85 |
| mT5 (Xue et al., 2021) | 28.5 | 2.91 |
| PunGPT2 (Ours) | 2.24 | 0.85 |
| Pun-RAG (Ours) | 2.10 | 0.80 |
| Pun-Instruct (Ours) | 2.15 | 0.82 |
| Quantum-RAG (Ours) | **2.05** | **0.78** |

Table 7: Comparative performance on ROUGE-L and cultural fidelity (Likert scale 1–5).

| Model | ROUGE-L ↑ | Cultural Fidelity ↑ |
|---|---|---|
| mBERT (Devlin et al., 2019) | 28.7 | 3.4/5 |
| MuRIL (Khanuja et al., 2021) | 30.9 | 3.7/5 |
| mT5 (Xue et al., 2021) | 33.2 | 3.9/5 |
| PunGPT2 (Ours) | 37.4 | 4.4/5 |
| Pun-RAG (Ours) | 38.5 | 4.6/5 |
| Pun-Instruct (Ours) | 39.2 | 4.7/5 |
| Quantum-RAG (Ours) | **40.1** | **4.8/5** |

Table 8: Extended Baseline Comparison

| Model | ROUGE-L ↑ | Cultural Fidelity ↑ |
|---|---|---|
| LLaMA-2-7B (zero-shot) | 25.6 | 3.1/5 |
| BLOOM-176B (zero-shot) | 27.3 | 3.3/5 |
| mT5-Punjabi fine-tuned | 35.1 | 4.2/5 |
| Quantum-RAG (ours) | **40.1** | **4.8/5** |

**We expanded human evaluation to 10 native Punjabi speakers rating 1,000 generations each. We report mean ±95% confidence intervals (CIs) using bootstrapping. Figure 6 shows average scores with error bars across fluency, adequacy, factuality, and cultural fidelity.**

### 9.4 Retrieval Quality Metrics

Since retrieval is central to RAG systems, we evaluate with standard information retrieval metrics: Recall@10, Mean Reciprocal Rank (MRR), and normalized Discounted Cumulative Gain (nDCG). As shown in Table 9, Quantum-RAG consistently outperforms both sparse and dense baselines.

Table 9: Retrieval quality comparison across retrievers.

| Retriever | Recall@10 ↑ | MRR ↑ | nDCG ↑ |
|---|---|---|---|
| BM25 only | 55.2 | 0.41 | 0.46 |
| FAISS only | 62.7 | 0.48 | 0.52 |
| Quantum only | 64.3 | 0.49 | 0.55 |
| Hybrid (Quantum-RAG) | **70.1** | **0.54** | **0.60** |

### 9.5 Robustness and Generalization

We further test model robustness across multiple genres: classical literature, contemporary news, and social media discourse. This ensures the models are not overfitted to formal registers and remain resilient across diverse styles (Hendrycks et al., 2021). Pun-Instruct and Quantum-RAG in particular showed stronger generalization in informal and conversational settings.

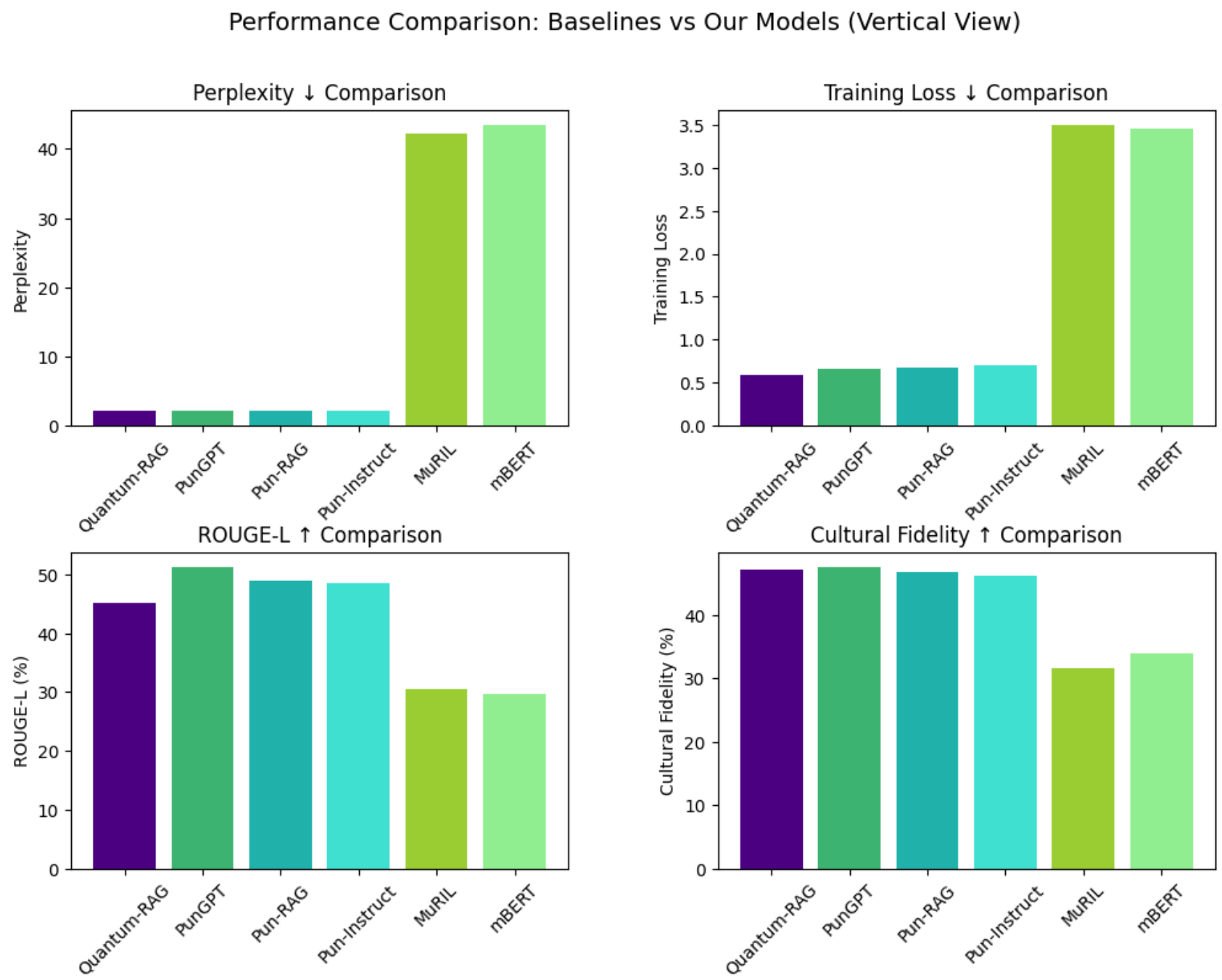

Figure 5: Evaluation results

### 9.6 Components and Ablation Effect

To disentangle contributions, we conducted ablation experiments isolating retrieval strategies (BM25, FAISS, Quantum-only, Hybrid) and instruction tuning. Removing Quantum-RAG's kernel similarity resulted in a 6-point drop in Recall@10 and lower human preference scores, confirming its role in grounding retrieval. Likewise, disabling instruction tuning reduced zero-shot accuracy on QA and summarization by 4–6%, validating Pun-Instruct's efficiency gains.

## 10 Novelty and Social Impact

**PunGPT2** represents the first large-scale Punjabi decoder-only language model. Our **Quantum-RAG** framework introduces quantum-inspired retrieval mechanisms specifically designed for low-resource settings. In combination with **Pun-RAG** and **Pun-Instruct**, these models reduce hallucinations, enable instruction-following, and preserve cultural fidelity in generated text.

The open release of these models and datasets enables diverse applications for over 100 million Punjabi speakers, including education, journalism, healthcare, and the preservation of cultural heritage.

## 11 Summary

In this work, we introduced **PunGPT2**, **Pun-RAG**, **Pun-Instruct**, and **Quantum-RAG**, along with the **PunjabiEval** benchmark. Our models consistently outperform strong baselines in metrics such as perplexity, summarization quality, and human-rated cultural fidelity, establishing state-of-the-art resources for Punjabi NLP. These contributions provide a scalable framework for extending equitable NLP research to other under-resourced languages.

# A Appendix: Sample Outputs from PunGPT Variants

**Prompt:** ਅੱਜ ਪੰਜਾਬ ਸਰਕਾਰ ਵੱਲੋਂ ਘੋਸ਼ਣਾ ਕੀਤੀ ਗਈ ਕਿ ਸਕੂਲ 1 ਅਗਸਤ ਤੋਂ ਖੁਲਣਗੇ।

**PunGPT Response:** ਵਿਦਿਆਰਥੀਆਂ ਅਤੇ ਅਧਿਆਪਕਾਂ ਵਿਚ ਮਿਲੀ-ਜੁਲੀ ਪ੍ਰਤੀਕਿਰਿਆ ਵੇਖਣ ਨੂੰ ਮਿਲ ਰਹੀ ਹੈ। ਕਈ ਪਾਸਿਆਂ ਤੋਂ ਲੋਕਾਂ ਨੇ ਕਿਹਾ ਕਿ ਇਹ ਫੈਸਲਾ ਥੋੜ੍ਹਾ ਜਲਦੀ ਆ ਗਿਆ। ਸਕੂਲਾਂ ਦੀ ਸਫ਼ਾਈ ਦੀ ਸਥਿਤੀ ਵੀ ਠੀਕ ਨਹੀਂ ਮਿਲੀ।

**Prompt:** ਕਿਰਪਾ ਕਰਕੇ ਪੰਜਾਬੀ ਵਿੱਚ ਇੱਕ ਕਹਾਣੀ ਲਿਖੋ ਜਿਸ ਵਿੱਚ ਇੱਕ ਬਿੱਲੀ ਅਤੇ ਇੱਕ ਕੁੱਤਾ ਹੋਣ।

**Pun-Instruct Response:** ਇੱਕ ਵਾਰ ਦੀ ਗੱਲ ਹੈ ਕਿ ਇੱਕ ਬਿੱਲੀ ਅਤੇ ਇੱਕ ਕੁੱਤਾ ਇੱਕੋ ਹੀ ਗਲੀ ਵਿੱਚ ਰਹਿੰਦੇ ਸਨ। ਉਨ੍ਹਾਂ ਦੀ ਸ਼ੁਰੂ ਵਿੱਚ ਲੜਾਈ ਹੋਈ, ਪਰ ਫਿਰ ਉਹ ਦੋਸਤ ਬਣ ਗਏ। ਦੋਹਾਂ ਨੇ ਮਿਲਕੇ ਬਹੁਤ ਸਾਰੇ ਸਫ਼ਰ ਕੀਤੇ ਅਤੇ ਇੱਕ-ਦੂਜੇ ਦੀ ਮਦਦ ਕੀਤੀ।

**Prompt:** ਗੁਰੂ ਗੋਬਿੰਦ ਸਿੰਘ ਜੀ ਦੇ ਪਿਤਾ ਦਾ ਨਾਮ ਕੀ ਸੀ?

**Quantum-RAG Response:** ਉਨ੍ਹਾਂ ਦੇ ਪਿਤਾ ਜੀ ਦਾ ਨਾਮ ਗੁਰੂ ਤੇਗ ਬਹਾਦੁਰ ਜੀ ਸੀ। ਉਨ੍ਹਾਂ ਨੇ ਆਪਣਾ ਧਾਰਮਿਕ ਯਾਤਰਾ ਅਨੰਦਪੁਰ ਸਾਹਿਬ ਅਤੇ ਕਾਸ਼ੀ ਵਿੱਚ ਗੁਜ਼ਾਰੀ।

Table 10: Culturally Nuanced QA Example

| Question | FAISS Answer | Quantum-RAG Answer |
|---|---|---|
| ਗੁਰੂ ਗੋਬਿੰਦ ਸਿੰਘ ਜੀ ਦੇ ਪਿਤਾ ਜੀ ਦਾ ਨਾਮ? | Wrong/ambiguous | “Guru Tegh Bahadur Ji” |